\documentclass[sigconf, nonacm]{acmart}
\usepackage{xspace}
\providecommand{\eg}{e.g.\xspace}
\providecommand{\ie}{i.e.\xspace}

\AtBeginDocument{%
  }

\setcopyright{acmlicensed}
\copyrightyear{2018}
\acmYear{2018}
\acmDOI{XXXXXXX.XXXXXXX}
\acmConference[Conference acronym 'XX]{Make sure to enter the correct
  conference title from your rights confirmation email}{June 03--05,
  2018}{Woodstock, NY}

\acmISBN{978-1-4503-XXXX-X/2018/06}

\acmSubmissionID{9177}

\usepackage{algorithm}
\usepackage{algorithmic}
\usepackage{colortbl}
\usepackage{bm}
\definecolor{Gray}{gray}{0.93}
\usepackage{newfloat}
\usepackage{listings}
\usepackage{multirow}
\usepackage{graphicx} 
\usepackage{subcaption}
\definecolor{mgreen}{rgb}{0.19,0.80,0.19}
\begin{document}

\title{HydraPrompt: An Adaptive and Asymmetric Framework of Vision-Language Models for Synthetic Image Detection}

\author{
Senyuan Shi\textsuperscript{1*} 
\quad
Hao Tan\textsuperscript{2 4}\textsuperscript{*} 
\quad
Zichang Tan\textsuperscript{3} \quad
Shuhan Feng\textsuperscript{1} \quad \\
Ajian Liu\textsuperscript{4} \quad
Sergio Escalera\textsuperscript{5} \quad
Jun Wan\textsuperscript{4}\textsuperscript{\S} \quad
}

\affiliation{
\vspace{2pt}
\textsuperscript{1}Beijing University of Posts and Telecommunications \country{}\\
\textsuperscript{2}School of Advanced Interdisciplinary Sciences (SAIS), University of Chinese Academy of Sciences \\
\textsuperscript{3}Shenzhen Institute of Advanced Technology (SIAT), Chinese Academy of Sciences \\
\textsuperscript{4}MAIS, Institute of Automation, Chinese Academy of Sciences \\
\textsuperscript{5}University of Barcelona
\textsuperscript{*} Equal contribution, \textsuperscript{\S} Corresponding author
}

\email{
442538552@bupt.edu.cn, {tanhao2023, jun.wan}@ia.ac.cn
}

\renewcommand{\shortauthors}{Shi et al.}

\begin{abstract}
The rapid evolution of generative models has precipitated a proliferation of fabricated content, posing significant challenges to existing Synthetic Image Detection (SID) methods. Capitalizing on advancements in vision-language models (\eg, CLIP), recent attempts have leveraged learnable textual prompts to identify synthetic images. However, they still leverage static prompt as a fixed boundary for real and fake images, failing to adapt to the varying types of forgery that emerge during inference.
To overcome this issue, we propose \textbf{HydraPrompt}, an asymmetric prompting framework that dynamically adjusts the category centers by aligning with fine-grained visual cues.
Specifically, we propose an Asymmetric Prompt Adapter (\textbf{APA}): (1) for \textit{authentic category}, we introduce a single set of prompts to capture the consistent representative patterns, which serves as a unified anchor for real content.
While (2) for \textit{fake category}, we construct sample-adaptive prompts that specialize in capturing diverse cues from different samples, enabling adaptive modeling of forgery image variations.
To increase pronounced discriminability within different synthetic images, we further introduce a Conditional Supervised Contrastive (\textbf{CSC}) objective, which compacts the authentic representations while capturing fine-grained forgery clues. 
Extensive experiments on popular SID benchmarks demonstrate the state-of-the-art performance of our framework.
\end{abstract}

\maketitle



\begin{figure}[t]
  \includegraphics[width=0.47\textwidth]{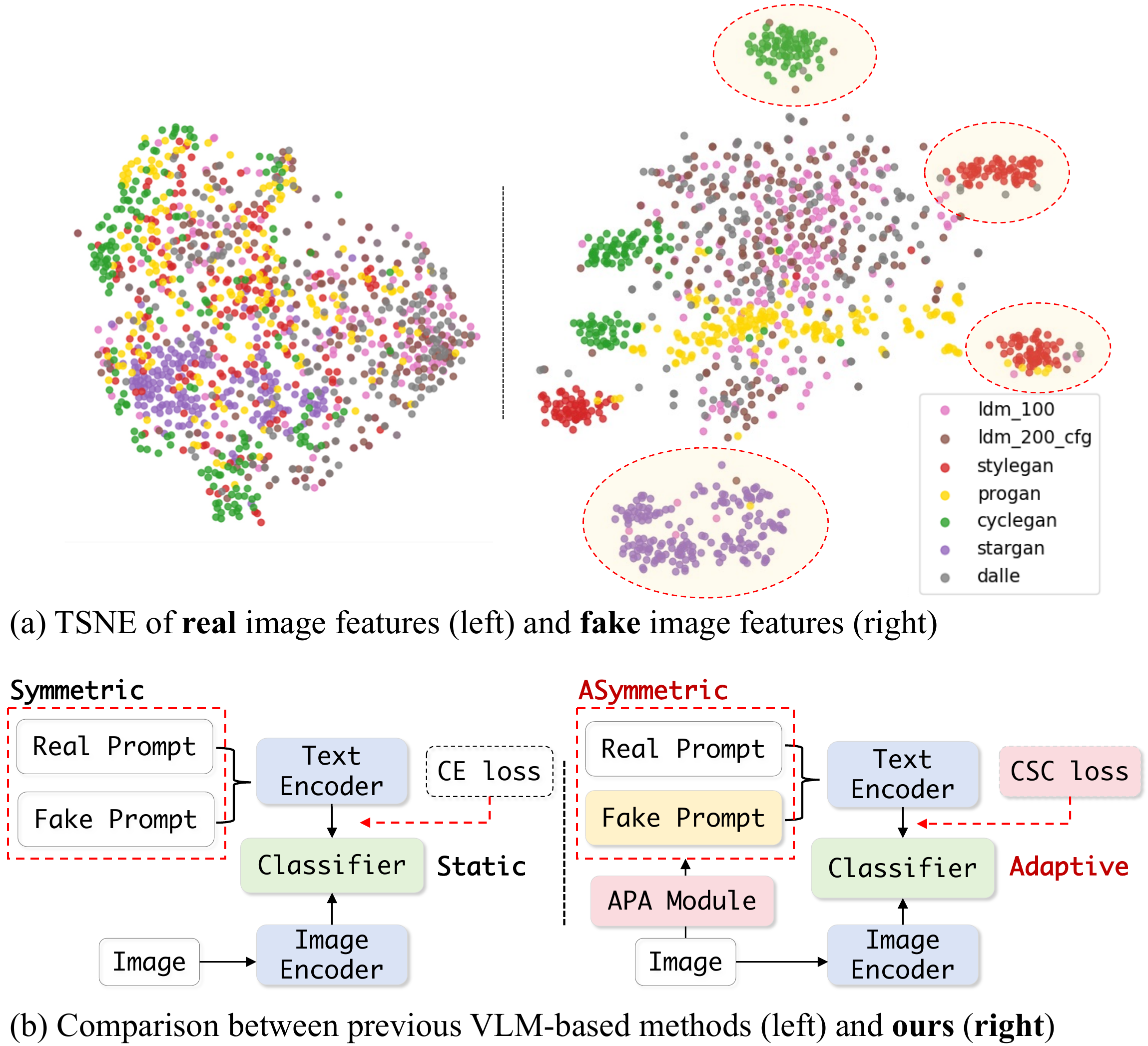} 
  \caption{
  (a) TSNE \cite{maaten2008visualizing} visualizations of real and fake images on UniversalFakeDetect \cite{ojha2023towards} benchmark. We selected seven subsets from the benchmark. The features are extracted from shallow layers of CLIP ViT-L/14. Real images share a unified distribution while fake images exhibit scattered patterns.
  (b) Comparison of previous VLM-based strategy and our HydraPrompt.
  Existing approaches adopt symmetric prompts that produce static category centers.
  In contrast, HydraPrompt introduces \textit{asymmetric} prompts design to achieve adaptive category centers.
  } 
  \label{figure:observation} 
  \vspace{-10pt}
\end{figure}

\begin{figure*}[t]
  \includegraphics[width=1\textwidth]{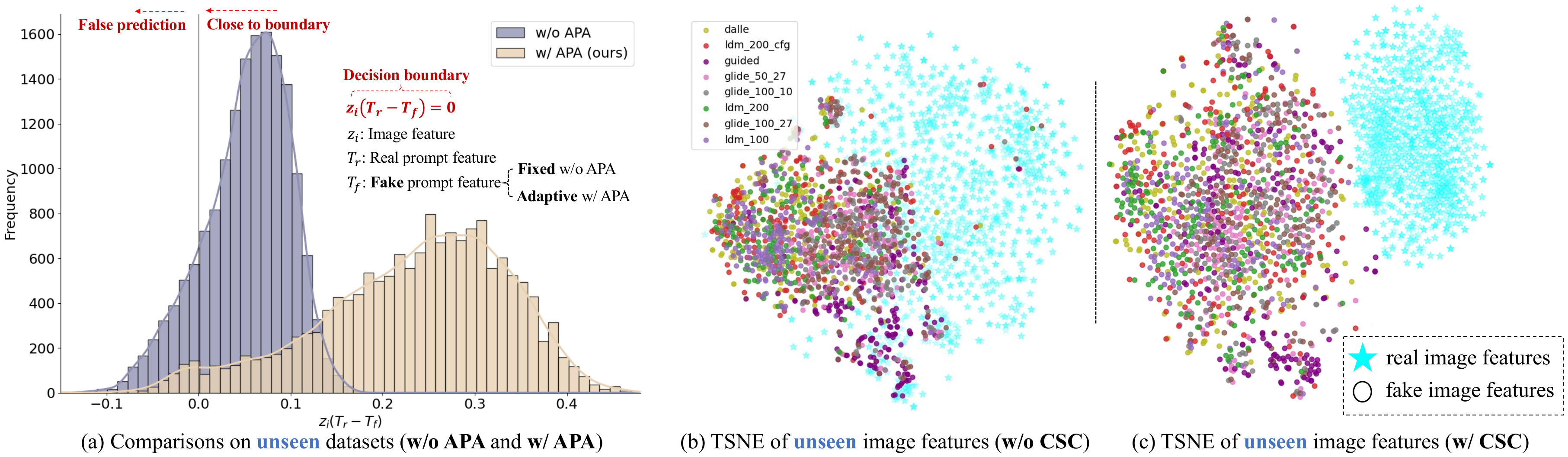} 
  \caption{Analyses on the proposed APA and CSC.
  \textbf{(a) Effectiveness of APA}: $\bm{z}_i\bm{T}_r$ and $\bm{z}_i\bm{T}_f$ are compared for classification during inference, thus $\bm{z}_i(\bm{T}_r-\bm{T}_f)=0$ is the decision boundary.
  The values $\bm{z}_i(\bm{T}_r-\bm{T}_f)$ for fake category are inverted so that positive values indicate correct predictions, and a value close to zero signifies proximity to the boundary (\ie, ambiguous cases).
  The results reveal that previous strategy is error-prone on unseen data.
  While the proposed APA constructs adaptive category centers, which increases the margin, leading to improved robustness.
  \textbf{(b) and (c) Effectiveness of CSC}: we compare the learned image features on unseen subsets from UniversalFakeDetect \cite{ojha2023towards} benchmark.
  Without CSC, the features of real and fake images exhibit obvious overlap.
  By contrast, CSC compacts the real representations and prevents the unseen forgeries from collapsing into a narrow region, showing better separability in OOD scenarios.}
  \label{figure:modules_res} 
\end{figure*}

\section{Introduction}
Recent years have witnessed the advancement of generative models, such as GANs \cite{gao2019progan, Karras_2020_CVPR} and LDMs \cite{Rombach_2022_CVPR}. These models enable the creation of hyper-realistic synthetic images, thus raising wide concerns of potential abuse and privacy threats. In response to these security concerns and privacy threats, a variety of detection methods targeting AI-generated images (AIGIs) have been proposed.

Most existing studies in AIGIs detection \cite{ojha2023towards, yan2024transcending, tan2024rethinking, nguyen2024laa,fu2025exploring, yan2024sanity, yan2024orthogonal, yang2025all, qian2020thinking, tan2024frequency, zhou2024freqblender, kashiani2025freqdebias} typically approach it as a symmetric binary classification task, akin to the ``cat versus dog'' problem.
This strategy proves effective in traditional scenarios since the categories exhibit well-defined differences in visual semantics. 
However, AIGIs tasks focus on fine-grained details rather than high-level visual semantics \cite{koutlis2024leveraging}.
In this regard, 
\textit{authentic images}, despite their diversity, possess a consistent pattern governed by the unified physics of real-world imaging, whereas \textit{synthetic images} contain a wide range of artificial patterns due to variations in generative models (\eg, GANs, LDMs) and forgery traces (\eg, structural or physical or even pixel flaws).

Specifically, we visualize the shallow layers image features from pre-trained encoder (Figure~\ref{figure:observation} (a)), as we hypothesize that the fine-grained details reside in the shallow features of the visual encoder \cite{koutlis2024leveraging}. 
It can be observed that there is a pronounced distribution shift when confronted with forgery data from different sources. 
In contrast, real images share a unified distribution, which reflects that there exists an \textit{asymmetric} phenomenon in the AIGIs detection task. 
Consequently, traditional symmetric strategy raises two crucial limitations:
\textbf{(1)} a mismatch between the fixed classification boundary and the diverse, nuanced visual cues inherent in manipulated content, leading to failures for out-of-distribution (OOD) forgeries.
\textbf{(2)} The reliance on fixed classification boundary restricts the model's ability to adapt to varying types of forgeries, prone to overfitting seen manipulation patterns (as in Figure~\ref{figure:modules_res} (b)).

Building on the above investigations, we propose \textbf{HydraPrompt}, a conceptually simple yet effective framework based on Vision-Language Models (VLMs), which mitigates the above issues by constructing asymmetric category centers for real and fake classes, as shown in Figure~\ref{figure:observation} (b).

To address the first issue, we introduce an \textbf{A}symmetric \textbf{P}rompt \textbf{A}dapter (APA).
Unlike existing adapters \cite{tan2024frequency, liu2024forgery} for SID tasks, which rely on a fixed set of textual prompts, we introduce an asymmetric prompt modeling.
Specifically, for real category we employ a single set of prompts to capture the consistent patterns of real-world content.
While for the fake category, we introduce sample-adaptive prompts based on fine-grained visual cues, aiming to adapt to diverse clues of forged images.
To capture the fine-grained image clues, we explore different methods such as LDR~\cite{chen2025ldr} that uses hand-drafted features, and RINE~\cite{koutlis2024leveraging} that leverages ensemble of spatial features.
We empirically found that the shallow-layer image features coupled with a simple learnable block, can effectively model diverse forged images.
As validated in Figure~\ref{figure:modules_res} (a), previous strategy is error-prone on unseen data, while our proposed APA greatly improves the classification margin through adaptive modeling of decision boundary, showing great robustness to unseen forgeries.

To mitigate the second issue, we propose \textbf{C}onditional \textbf{S}upervised \textbf{C}ontrastive (CSC) objective.
Specifically, (1) for the image branch, we pull real image features closer toward the unified anchor, 
while dispersing fake image features to prevent the model from collapsing into seen forgeries.
(2) For the text branch, we explicitly separate the text features to preserve their distinctiveness.
Additionally, inspired by \cite{wang2020understanding}, we introduce an alignment constraint as a regularizer to align cross-modal representations progressively.
As shown in Figure~\ref{figure:modules_res} (b) and (c), CSC effectively compacts the representations of real images while dispersing those of fake images, offering better distinctiveness to unseen forgeries.
During inference, we compare the cosine similarities between image features and two category centers (\ie, unified real category center and sample-adaptive fake category center) to determine the authenticity.

To sum up, our main contributions are:
\begin{itemize}
    \item 
    We introduce HydraPrompt, a conceptually simple yet effective framework that dynamically constructs asymmetric prompts to achieve adaptive category centers,
    achieving substantial gains on popular SID benchmarks.
    
    \item We propose \textbf{A}symmetric \textbf{P}rompt \textbf{A}dapter (APA), 
    which leverages fine-grained visual cues to construct adaptive category centers, enhancing the classification margin statistically on unseen forgeries.
    
    \item We propose \textbf{C}onditional \textbf{S}upervised \textbf{C}ontrastive (CSC) Objective, which facilitates alignment between visual features and asymmetric textual prompts, enhancing the adaptation ability in OOD scenarios.
    
\end{itemize}

\section{Related Work}
\label{sec:Related Work}

\subsection{Generalizable Synthetic Image Detection}
\label{ssec:synthetic images detection}
The evolution of generative models~\cite{esser2024scaling, tian2024visual, ye2025dreamid} has posed significant security risks to society.
Synthetic image detection, which aims to distinguish AI-generated images from authentic images, has become a heated topic.
Early studies focus on finding generalizable \textit{features} in different domains, such as spatial domain~\cite{ojha2023towards, yan2024transcending, tan2024rethinking, nguyen2024laa,fu2025exploring, yan2024sanity, yan2024orthogonal, yang2025all}, frequency domain~\cite{qian2020thinking, tan2024frequency, zhou2024freqblender, kashiani2025freqdebias}, time-domain~\cite{gu2021spatiotemporal, gu2022hierarchical, gu2022delving, yan2025generalizing} and even model gradients~\cite{tan2023learning}.
While the tailored feature extractions fail to generalize well in practical scenarios~\cite{yang2025d}.
Some methods discover that the reconstruction error is a generalizable indicator for synthetic images, hence employing VAE~\cite{ricker2024aeroblade, chu2025fire} or Denoising U-Net~\cite{chen2024drct, wang2023dire, cazenavette2024fakeinversion} for detection. However, the performance is conditioned on the specific type of VAE and is hard to generalize to GAN-generated images.
Some methods aim to mitigate bias in aligned datasets. For instance, \cite{guillaro2025bias} \cite{rajan2024aligned} align content bias between authentic and synthetic categories through VAE reconstruction, while \cite{chen2025dual} \cite{corvi2025seeing} mitigate frequency bias by further aligning high-frequency information. However, a large amount of bias remains unexplored. Such specific training approaches can make the model prone to overfitting to unseen distributional bias. 
More recently, some methods~\cite{koutlis2024leveraging, tsai2024understanding, he2024rigid, yan2024orthogonal} take advantage of pre-trained semantics to greatly boost the performance.
For instance, Effort~\cite{yan2024orthogonal} utilizes the well-established semantic knowledge to enhance the generalization of synthetic image detection, achieving impressive performance.
Although these methods utilize advanced feature extraction strategies, the potential of dynamic category centers is still underexplored.

\begin{figure*}[t]
  \centering 
  \includegraphics[width=1.0\textwidth]{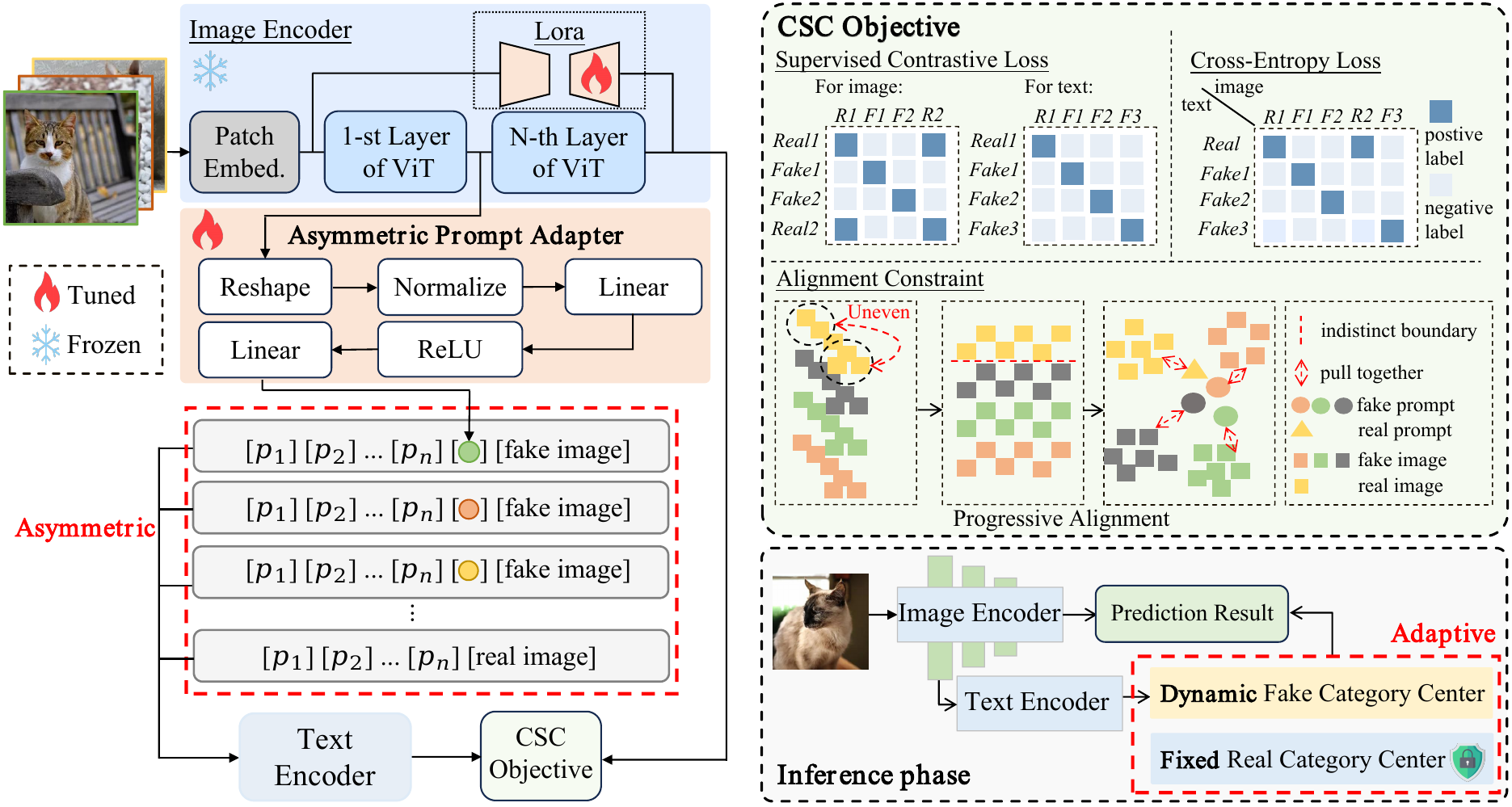} 
  \caption{\textbf{Overview of the proposed HydraPrompt framework.} During training phase, we design sample-adaptive prompts for fake image features, while using a single set of learnable prompts for real image features. ${P_{n}}$ denotes the learnable prompts. The CSC objective is employed to align both intra-modal and cross-modal embedding space. In inference phase, we obtain real logit by calculating the cosine similarity between image features and fixed real text prompts, and the fake logit by calculating the cosine similarity between image features and adaptive prompts. (1)~\textbf{Asymmetric}: the distinct design of textual embeddings for different categories. (2)~\textbf{Adaptive}: fine-grained visual cues of test image are extracted as part of sample-adaptive prompts, enabling the achievement of adaptive category centers.  } 
  \label{fig:method} 
\end{figure*}

\subsection{VLMs for Synthetic Image Detection}
With the developments of Vision-Language Models (VLMs)~\cite{radford2021learning, liu2023visual}, many attempts tend to ground their capacities into the field of synthetic image detection.
For instance, some methods~\cite{chen2024x2, xu2024fakeshield, huang2025sida, zhou2025aigi} adopt large VLMs for explainable detection, achieving promising results and transparency.
Other methods take advantage of CLIP~\cite{liu2023visual}.
For example, FatFormer~\cite{liu2024forgery} integrates a forgery-aware adapter and dedicated language alignments into CLIP, yielding remarkable performance.
C2P-CLIP~\cite{tan2025c2p} introduces category common prompts to align the cross-modal embeddings in the forgery space.
Sun et al.~\cite{sun2025towards} construct precise forgery explanations to release the potential of CLIP and large VLMs.
Lin et al.~\cite{lin2025standing} and Forensics adapter~\cite{cui2025forensics} employ learnable prompts to transfer cross-modal knowledge into facial forgery detection.
While achieving significant improvements, they still rely on static category centers, neglecting the adaptation capacities of the textual encoder.
In this paper, we present an asymmetric cross-modal framework, leveraging fine-grained visual cues to construct dynamic category centers for generalizable forgery detection.

\subsection{Prompt Learning for VLMs}
Large-scale pre-trained Vision-Language models (VLMs), \eg, CLIP~\cite{radford2021learning} and ALIGN~\cite{jia2021scaling} have demonstrated remarkable generalization across various downstream tasks~\cite{wysoczanska2024clip, tan2025recover}, ~\cite{zhu2025uniemo, zhu2026H-GAR, zhu2025emosym}, \cite{tan2025veritas, tan2026videoveritas}.
How to efficiently transfer these VLMs into specific domains remains a heated topic.
Various methods have been proposed, \eg, prompt learning~\cite{zhou2022learning}, adapter tuning~\cite{gao2024clip} and LoRA~\cite{hu2022lora}.
Among them, the prompt learning technique is widely adopted for its effectiveness and limited computational overhead.
Different from hand-crafted textual templates, prompt learning introduces trainable prompt embeddings to facilitate the learning of domain knowledge.
For instance, 
CoOp~\cite{zhou2022learning} and CoCoOp~\cite{zhou2022conditional} introduce learnable textual prompts to transfer CLIP into classification task, yielding impressive few-shot performance.
The follow-up works~\cite{zhu2023prompt, khattak2023maple} further improve the generalization abilities either through gradient correction~\cite{zhu2023prompt}, multi-modal prompting~\cite{khattak2023maple} or ensemble learning~\cite{BeyondSoloStrength}.
In this paper, we inherit the merits of learnable prompts, while further incorporating fine-grained visual cues as cross-modal prompts to construct an inference-adaptive framework.

\section{Methodology}

In this section, we present HydraPrompt, an adaptive and Asymmetric Framework for Synthetic Image Detection. 
We first define some basic notations and review the classification strategy of CLIP~\cite{radford2021learning}.
Then we give a detailed demonstration of our proposed method, including Asymmetric Prompt Adapter (\textbf{APA}) Module, Conditional Supervised Contrastive (\textbf{CSC}) Objective and inference procedure.

\subsection{Preliminaries}
\noindent \textbf{Notations.}
Suppose the dataset contains $W$ images, denoted as $\{\bm{I}_i, \bm{y}_i\}_{i=1}^W$.
Each image is annotated with a binary label $\bm{y}_i$ indicating its authenticity.
$\bm{y}_i=0$ represents the image is real and $\bm{y}_i=1$ indicates the image is tampered or entirely generated by AI.
Our method is constructed on the CLIP.
We denote the image encoder and the text encoder pre-trained from CLIP as $\mathcal{V}(\cdot)$ and $\mathcal{T}(\cdot)$, respectively.

\noindent \textbf{Classification Strategy of CLIP.}
Given an input image and candidate labels, 
the classification is performed by comparing the cosine similarities of the image feature and textual embeddings,
where textual embeddings are actually serving as the category centers.
Different from traditional vision foundation networks, the semantic knowledge embedded from text encoders are well-aligned with the image feature space. However, the textual embeddings remain static during inference, as in existing SID methods~\cite{cui2025forensics,guo2025rethinking}, extracted either by hand crafted templates or the learnable prompts, failing to fully release the potential of the text encoder.

\subsection{Overall Architecture}
\label{sec:Overall Architecture}
With the primary objective of establishing adaptive category centers to ensure that the model maintains strong generalization on unseen data, we introduce two core designs: APA module and CSC objective.
As shown in Figure~\ref{fig:method},
(1) For \textbf{APA Module}, we extract the shallow-layer features from the image encoder and integrate learnable prompts to efficiently capture sample-adaptive prompts.
(2) For \textbf{CSC Objective}, our framework employs two loss functions, where the supervised contrastive loss aligns the intra-modal embeddings comprised both image and text branches, while the alignment constraint handles the cross-modal alignment.

\subsection{APA Module}
\noindent \textbf{Fine-grained Visual Cues.}
AIGIs tasks focus on fine-grained visual cues rather than high-level visual semantics. Therefore, we conclude that shallow layers features from image encoder can serve as effective fine-grained visual cues.
\noindent Specifically, given an input image, we first extract patch embeddings to obtain $\bm{Z} \in \mathbb{R}^{N \times D}$, where $N$ denotes the number of patches and $D$ represents the embedding dimension. The image encoder processes these embeddings through a sequence of transformer blocks. Let $\mathcal{V}_i$ denote the $i$-th transformer block in the image encoder. The image features after the first block of image encoder are obtained as: $\bar{\bm{Z}} = \mathcal{V}_1(\bm{Z})$. Then we apply average pooling over these image features, yielding an aggregate representation as:
\begin{equation}
\label{eq:1234}
\hat{\bm{Z}} = \frac{1}{N} \sum_{i=1}^{N} {\bar{\bm{z}}}_i,
\quad {\bar{\bm{z}}}_i \in \mathbb{R}^{D}.
\end{equation}
Then, we employ a lightweight learnable adapter to project the extracted features $\hat{\bm{Z}}$. The process can be formulated as:
\begin{equation}
\label{eq:123}
\tilde{\bm{Z}} = \bm{W}_2^\mathsf{T} \big( \text{ReLU}( \bm{W}_1^\mathsf{T} \hat{\bm{Z}} + \bm{b}_1 ) \big) + \bm{b}_2,
\end{equation}
where $\bm{W}_1 \in \mathbb{R}^{D \times D^{\prime}}$ and
$\bm{W}_2 \in \mathbb{R}^{D^{\prime} \times D}$ are trainable weights with $D^{\prime}<D$.
$\bm{b}_1$ and $\bm{b}_2$ are bias terms.
$\tilde{\bm{Z}}$ is the modulated feature containing fine-grained visual cues.

\noindent \textbf{Construction of adaptive category centers.}
To address the mismatch between fixed classification boundary and diverse, nuanced visual cues in manipulated content, we introduce sample-adaptive prompts based on the fine-grained visual cues. Specifically, we construct \textbf{two} sets of textual embeddings for each input image. 
Firstly, we employ learnable prompts $\bm{P}_r\in \mathbb{R}^{M\times D}$ to extract a consistent textual embedding, where $M$ denotes the number of prompts.
This remains static during inference, serving as a stable anchor for authentic images.
Secondly, we take fine-grained visual cues $\tilde{\bm{Z}}$ as a cross-modal prompt, together with another set of learnable prompts $\bm{P}_f\in \mathbb{R}^{M\times D}$ to construct sample-adaptive textual embeddings. The above process can be written as:
\begin{equation}
\begin{aligned}
    \bm{T}_{r} &= \mathcal{T}([\bm{P}_r, \bm{C}_r]), \\
    \bm{T}_{f} &= \mathcal{T}([\bm{P}_f, \tilde{\bm{Z}}, \bm{C}_f]),
\end{aligned}
\label{eq:textfea}
\end{equation}
where $\bm{C}_r$ is the context vector obtained from the text ``\texttt{A real image}'' and $\bm{C}_f$ is the context vector obtained from ``\texttt{A fake image}'', serving as the common anchors for the corresponding categories.
$\bm{T}_{r}$ is the unified textual embeddings for authentic category and $\bm{T}_{f}$ is the sample-adaptive textual embeddings for synthetic category. $[\cdot]$ denotes concatenation operation.
Conditioning on fine-grained visual cues, $\bm{T}_{f}$ is capable of constructing sample-adaptive prompts, effectively adapting the category centers.

\subsection{CSC Objective}

\noindent \textbf{Supervised Contrastive Objective.}
Based on the asymmetric architecture, we encourage the features of the real class to cluster together, while the features of the fake class remain dispersed. The process can be formulated as:

\begin{equation}
\label{eq:supcon}
\mathcal{L}_{sc}^{I}=\frac{1}{B}\!\sum_{i=1}^{B}\!\frac{-1}{\sum_{j=1}^{B}\! \mathbf{M}_{ij}} \log\!\frac{\sum_{j=1}^{B}\!\mathbf{M}_{ij}e^{\left(\bm{z}_{i}\cdot \bm{z}_{j}/\tau\right)\!}}{\sum_{k=1}^{B}\!e^{\left(\bm{z}_{i}\cdot \bm{z}_{k}/\tau\right)}},
\end{equation} 

\begin{equation}
\label{eq:Mij}
\mathbf{M}_{i j}=\left\{\begin{array}{ll}
1, & \text{if } \bm{y}_{i}=0 \text{ and } \bm{y}_{j}=0, \\
1, & \text{if } \bm{y}_{i}=1 \text{ and } i=j, \\
0, & \text{otherwise},
\end{array}\right. 
\end{equation} 

\begin{equation}
\label{eq:Lsc}
\mathcal{L}{sc} = \frac{1}{2}(\mathcal{L}_{sc}^{I} + \mathcal{L}_{sc}^{T}),
\end{equation}

\noindent
where $B$ denotes the batch size, and $\mathbf{z}_i$ represents the normalized feature of the $i$-th image in the batch. $\mathcal{L}_{sc}^{I}$ and $\mathcal{L}_{sc}^{T}$ refer to the image and text branch, which share the analogous formula in the supervised contrastive objective.
The temperature parameter $\tau$ controls the sharpness of the similarity distribution.
The mask matrix $\mathbf{M} \in \{0, 1\}^{B \times B}$ defines positive pairs for the supervised contrastive objective. There are two strategies: (1) ``individual'', this strategy denotes that each feature is regarded as an individual by pushing the feature apart from one another. (2) ``cluster'', this strategy denotes all features are regarded as a single cluster by pulling them together. In our framework, we employ a ``cluster'' strategy for the authentic category and an ``individual'' strategy for the synthetic category as shown in Eq.~\ref{eq:Mij}.

\noindent \textbf{Alignment Constraint.}
To further enhance the construction of adaptive category centers, 
we incorporate an alignment constraint for cross-modal alignment.
Inspired by~\cite{wang2020understanding}, we formulate the alignment constraint as follows:

\begin{equation}
\label{eq:align}
\mathcal{L}_{\text{align}} = \frac{1}{B}\!\sum_{i=1}^{B}\!\frac{-1}{\sum_{j=1}^{B}\!\mathbf{M}_{ij}} 
\log\!\frac{\sum_{j=1}^{B}\!\mathbf{M}_{ij}e^{\left(\bm{z}_{i}\cdot \bm{t}_{j}^\top/\tau\right)\!}}{\sum_{k=1}^{B}\!e^{\left(\bm{z}_{i} \cdot\bm{t}_{k}^\top/\tau\right)}},
\end{equation}

\noindent
where $\bm{t}_{j}^\top$ denotes the normalized features of the $j$-th text in the batch. The mask matrix $\mathbf{M}$ is the same as Eq.~\ref{eq:Mij}. This term pulls positive pairs closer, aligning real representations and sample-adaptive fake representations progressively.

\noindent \textbf{Classification.}
Finally, we introduce a classification loss $\mathcal{L}_{\text{cls}}$,
which is a variant of standard cross-entropy loss adapted to asymmetric prompting framework.
Specifically, there are two output logits $\{o_r^i, o_f^i\}$ for the $i^{th}$ input.
The first one is computed from the static embedding $\bm{T}_r$ while the second is based on sample-adaptive embedding $\bm{T}_f^i$:
\begin{equation}
\begin{aligned}
    o_r^i = \bm{z}_i\cdot \bm{T}_r, o_f^i = \bm{z}_i\cdot \bm{T}_f^i.
\end{aligned}
\label{eq:logit}
\end{equation}
The classification loss can be calculated as:
\begin{equation}
\label{eq:cls}
\mathcal{L}_{\text{cls}}\!=\!-\!\!\sum_{i=1}^B\!\!\left(\!\bm{y}_i\log(\!\frac{e^{o_f^i}}{e^{o_r^i}\!+\!e^{o_f^i}})\!+\!(1\!\!-\!\bm{y}_i)\!\log(\!\frac{e^{o_r^i}}{e^{o_r^i}\!+\!e^{o_f^i}}\!)\!\!\right).
\end{equation}

\noindent \textbf{Final Objective.}
The overall loss function combines the classification loss, supervised contrastive loss, and the alignment constraint with their respective hyperparameters:
\begin{equation} 
\label{eq:overall}
\mathcal{L}_{\text{}} = \mathcal{L}_{\text{cls}} + \lambda_1 \mathcal{L}_{\text{sc}} 
+ \lambda_2 \mathcal{L}_{\text{align}},
\end{equation}
where ${\lambda_1}$ and $\lambda_2$ are coefficients that balance the importance of the corresponding regularization term.

\subsection{Inference}
During the inference stage,
there are two candidate centers for each input image, \ie, $\bm{T}_r$ and $\bm{T}_f$ calculated by Eq.~\ref{eq:textfea}.
Then we calculate their distance to the current image by using Eq.~\ref{eq:logit}.
By comparing the distances (\ie, $o_r^i$ and $o_f^i$), we can determine the authenticity of the current input image.

\begin{table*}[t] 
\centering
\footnotesize  
\renewcommand{\arraystretch}{1.0}
\newcolumntype{|}{!{\vline}}
\begin{tabular}{
    >{\raggedright\arraybackslash}p{2.5cm} | 
    >{\centering\arraybackslash}p{1.35cm} 
    >{\centering\arraybackslash}p{1.35cm}
    >{\centering\arraybackslash}p{1.35cm}
    >{\centering\arraybackslash}p{1.35cm}
    >{\centering\arraybackslash}p{1.35cm}
    >{\centering\arraybackslash}p{1.35cm}
    >{\centering\arraybackslash}p{1.35cm} |
    >{\centering\arraybackslash}p{1.35cm}
    >{\centering\arraybackslash}p{1.35cm}
}
\toprule
\textbf{Methods} 
& ProGAN
& StyleGAN
& BigGAN
& CycleGAN
& {GauGAN} 
& {StarGAN} 
& {Deepfake} 
& Mean \\
\midrule 
\midrule
CNNSpot \cite{wang2020cnn}  & 64.6~/~92.7 & 52.8~/~82.8 & 51.6~/~70.5 & 58.6~/~81.5 & 51.2~/~74.3 & 53.6~/~86.6 & 50.6~/~51.5 & 57.3~/~79.6 \\
Durall \cite{durall2020watch}  & 79.0~/~73.9 & 63.6~/~58.8 & 69.5~/~62.9 & 65.4~/~60.8 & 99.4~/~99.4 & 67.0~/~63.0 & 50.5~/~50.2 & 70.2~/~66.4 \\
Frank \cite{frank2020leveraging}  & 85.7~/~81.3 & 73.1~/~68.5 & 76.9~/~70.8 & 86.5~/~80.8 & 85.0~/~77.0 & 67.3~/~65.3 & 50.1~/~55.3 & 75.0~/~71.2 \\
F3Net \cite{qian2020thinking}  & 97.9~/~100 & 84.5~/~99.5 & 65.5~/~73.4 & 81.2~/~89.7 & 100~/~100 & 57.0~/~59.2 & 59.9~/~83.0 & 78.5~/~88.1 \\
LGrad \cite{tan2023learning}  & 99.8~/~100 & 94.8~/~99.7 & 82.5~/~92.4 & 85.9~/~94.7 & 99.7~/~99.9 & 73.7~/~83.2 & 60.6~/~67.8 & 86.2~/~92.2 \\
UnivFD \cite{ojha2023towards}  & 99.7~/~100 & 78.8~/~97.4 & 91.2~/~99.0 & 91.9~/~99.8 & 96.3~/~99.9 & 91.9~/~100 & 80.0~/~89.4 & 88.1~/~97.8 \\
FreqNet \cite{tan2024frequency}  & 97.9~/~99.6 & 97.6~/~99.9 & 90.5~/~96.1 & 95.8~/~99.5 & 90.2~/~99.7 & 93.4~/~98.6 & 97.4~/~99.9 & 94.7~/~99.0 \\
RINE \cite{koutlis2024leveraging}  & 100~/~100 & 99.3~/~100 & 99.6~/~99.9 &  88.9~/~99.4 & 99.5~/~100 & 99.8~/~100 & 80.6~/~97.9 & 95.4~/~99.6 \\
FatFormer \cite{liu2024forgery}  & 99.8~/~100 & 99.9~/~100 & 98.9~/~99.9 & 87.7~/~97.4 & 100~/~100 & 99.9~/~100 & 89.4~/~97.3 & 96.5~/~99.2 \\
C2P-CLIP \cite{tan2025c2p}  & 99.7~/~100 & 99.4~/~100 & 95.3~/~99.9 & 90.7~/~99.9 & 95.3~/~100 & 96.6~/~100 & 89.9~/~97.3 & 95.3~/~99.6 \\
Effort \cite{yan2024orthogonal}  & 100~/~100 & 95.1~/~100 & 99.6~/~99.9 & 99.9~/~97.7 & 99.6~/~100 & 100~/~100 & 87.6~/~98.9 & 97.4~/~99.5 \\
\midrule
\rowcolor{Gray}HydraPrompt (Ours)  & 99.9~/~100 & 96.9~/~100 & 98.8~/~99.8 & 98.2~/~99.9 & 98.4~/~100 & 100~/~100 & 93.7~/~97.5 & \textbf{98.0~/~99.6} \\
\bottomrule
\end{tabular}
\caption{Accuracy and average precision comparisons with state-of-the-art methods in \textbf{GANs} datasets on UniversalFakeDetect \cite{ojha2023towards} benchmark. We report the performance in the formulation of Acc / AP. The best results are highlighted in \textbf{bold}.}
\label{table:GAN datasets on UniFD}
\begin{tabular}{
    >{\raggedright\arraybackslash}p{2.4cm} | 
    >{\centering\arraybackslash}p{1.18cm} 
    >{\centering\arraybackslash}p{1.18cm}
    >{\centering\arraybackslash}p{1.18cm}
    >{\centering\arraybackslash}p{1.18cm}
    >{\centering\arraybackslash}p{1.18cm}
    >{\centering\arraybackslash}p{1.18cm}
    >{\centering\arraybackslash}p{1.18cm}
    >{\centering\arraybackslash}p{1.18cm} |
    >{\centering\arraybackslash}p{1.18cm}
}
\toprule
\textbf{Methods} 
& Guided
& DALL-E
& LDM 200 steps
& LDM 200 w/cfg
& LDM 100 steps
& Glide 100-27
& Glide 50-27
& Glide 100-10
& Mean \\
\midrule
\midrule
CNNSpot \cite{wang2020cnn} & 54.9~/~66.6 & 51.8~/~61.3 & 52.0~/~64.5 & 51.6~/~63.1 & 51.9~/~63.7 & 53.0~/~71.3 & 54.2~/~76.0 & 53.3~/~72.9 & 52.4~/~70.1 \\
Durall \cite{durall2020watch}  & 40.6~/~42.3 & 55.9~/~58.0 & 61.7~/~61.7 & 58.4~/~58.5 & 62.0~/~62.6 & 48.9~/~46.9 & 51.7~/~49.9 & 54.9~/~52.3 & 51.7~/~51.8 \\
Frank \cite{frank2020leveraging} & 53.4~/~52.5 & 57.0~/~62.5 & 56.4~/~50.9 & 56.5~/~52.1 & 56.6~/~51.3 & 50.4~/~40.8 & 52.0~/~42.3 & 53.6~/~44.3 & 53.2~/~50.2 \\
F3Net \cite{qian2020thinking}  & 69.2~/~70.8 & 71.6~/~79.9 & 73.4~/~83.3 & 80.7~/~89.1 & 74.1~/~84.0 & 87.0~/~94.5 & 88.5~/~95.4 & 88.3~/~95.4 & 80.6~/~89.2 \\
LGrad \cite{tan2023learning} & 86.6~/~100 & 88.5~/~97.3 & 94.2~/~99.1 & 95.9~/~99.2 & 94.8~/~99.2 & 87.4~/~95.1 & 90.7~/~95.1 & 89.4~/~94.9 & 89.4~/~97.7 \\
UnivFD \cite{ojha2023towards}  & 75.7~/~85.1 & 89.5~/~96.8 & 90.2~/~97.1 & 77.3~/~88.6 & 90.5~/~97.0 & 90.7~/~97.2 & 91.1~/~97.4 & 90.1~/~97.0 & 85.4~/~94.6 \\
FreqNet \cite{tan2024frequency}  & 86.7~/~96.3 & 59.1~/~77.8 & 84.6~/~96.1 & 99.6~/~100 & 65.6~/~62.3 & 85.7~/~99.8 & 97.4~/~99.8 & 88.2~/~96.4 & 83.4~/~91.1 \\
RINE \cite{koutlis2024leveraging} & 76.1~/~96.4 & 95.0~/~99.3 & 98.3~/~99.8 & 88.2~/~98.3 & 98.6~/~99.9 & 88.9~/~93.8 & 92.6~/~99.3 & 90.7~/~98.9 & 91.1~/~98.8 \\
FatFormer \cite{liu2024forgery} & 76.1~/~92.0 & 98.8~/~99.8 & 98.6~/~99.8 & 94.9~/~99.1 & 98.7~/~99.9 & 94.4~/~99.1 & 94.7~/~99.4 & 94.2~/~99.2 & 93.8~/~98.5 \\
C2P-CLIP \cite{tan2025c2p}  & 69.1~/~92.2 & 98.6~/~99.9 & 99.3~/~100 & 97.3~/~99.8 & 99.3~/~100 & 95.3~/~99.3 & 95.3~/~99.3  & 96.1~/~99.4 & 93.8~/~98.7 \\
Effort \cite{yan2024orthogonal} & 69.2~/~95.4 & 98.1~/~99.9 & 99.3~/~99.9 & 96.8~/~99.9 & 99.5~/~100 & 97.5~/~99.9 & 97.8~/~99.9 & 97.2~/~99.9 & 94.4~/~99.4 \\
\midrule
\rowcolor{Gray}HydraPrompt (Ours) & 89.5~/~97.6 & 98.4~/~99.9 & 99.5~/~100 & 97.3~/~99.9 & 99.6~/~100 & 93.7~/~99.4 & 95.4~/~99.6 & 93.7~/~99.7 & \textbf{95.9}~/~\textbf{99.5} \\
\hline
\end{tabular}
\caption{Accuracy and average precision comparisons with state-of-the-art methods in \textbf{diffusion model} datasets on UniversalFakeDetect \cite{ojha2023towards} benchmark. We report the performance in the formulation of Acc / AP. The best results are highlighted in \textbf{bold}.}
\label{table:DMs datasets on UniFD}
\end{table*}

\section{Experiments}

\subsection{Implementation Details}
\label{sec:Implementation Details}
We utilize CLIP ViT-L/14~\cite{radford2021learning} as the default vision-language model.
The input images are first resized to $256 \times 256$ resolutions, and then cropped into $224 \times 224$ resolutions.
No further data augmentations are performed during training phase.
$\lambda_1$, $\lambda_2$ are set to $1$ and $1.25$.
The model is trained for $10$ epochs with a learning rate of $4e-4$ and decays with cosine policy. To ensure a sufficient number of negative samples, we adopt the standard memory bank mechanism which stores and updates sample representations across training iterations. Therefore, the batch size can be increased to 1000 on a single RTX 4090 GPU. LoRA is integrated into the MLPs of all layers with rank 6 and alpha 6. The corresponding experiments are provided in the supplementary material.

\subsection{Datasets and Metrics}
We conducted experiments on three popular SID benchmarks: UniversalFakeDetect \cite{ojha2023towards}, Chameleon \cite{yan2024sanity}, and WildRF \cite{cavia2024real}. (1) For \textbf{UniversalFakeDetect} benchmark, we adhere to the protocol outlined in \cite{ojha2023towards} and utilize ProGAN as our training dataset, which includes 20 subsets of generated images. For evaluation, we follow the testing GANs datasets setting \cite{gao2019progan, chu2017cyclegan, karras2019style, brock2018large, choi2018stargan, park2019gaugan, rossler2019faceforensics++} and diffusion model datasets setting \cite{dhariwal2021diffusion,ramesh2021zero, rombach2022high, nichol2021glide} in FatFormer \cite{liu2024forgery}. Following existing work \cite{liu2024forgery}, we report both average precision (AP) and classification accuracy (Acc). For Acc, we set the classification threshold for each dataset at 0.5 to ensure fair comparison. (2) For \textbf{Chameleon} dataset, which contains widespread post-processing artifacts and is designed to evaluate model performance on challenging data, we adhere to the protocol outlined in \cite{yan2024sanity} and adopt two training settings: one using ProGAN \cite{gao2019progan} and the other using SD v1.4 \cite{ho2020denoising}. Following existing work \cite{yan2024sanity}, we report the Acc metric. (3) For \textbf{WildRF} dataset, which aims to address the difficulty of AIGIs detection in online environments, we adhere to the protocol outlined in \cite{cavia2024real}, and the test datasets include Reddit, Twitter, Facebook. We report the Acc metric.

\begin{table*}[t]
\centering
\footnotesize
\renewcommand{\arraystretch}{1.0}
\newcolumntype{|}{!{\vline}}
\vspace{-5pt}
\begin{tabular}{
    >{\raggedright\arraybackslash}p{1.7cm} | 
    >{\centering\arraybackslash}p{0.8cm}
    >{\centering\arraybackslash}p{0.8cm}
    >{\centering\arraybackslash}p{0.8cm}
    >{\centering\arraybackslash}p{0.8cm}
    >{\centering\arraybackslash}p{0.8cm}
    >{\centering\arraybackslash}p{0.8cm}
    >{\centering\arraybackslash}p{0.8cm}
    >{\centering\arraybackslash}p{0.85cm}
    >{\centering\arraybackslash}p{0.85cm}
    >{\centering\arraybackslash}p{0.95cm} |
    >{\centering\arraybackslash}p{2.25cm} 
}
\toprule
\textbf{Training Data} 
& \textbf{CNNSpot}
& \textbf{FreDect}
& \textbf{Fusing}
& \textbf{GramNet}
& \textbf{LNP}
& \textbf{UnivFD}
& \textbf{DIRE}
& \textbf{PatchCraft}
& \textbf{NPR}
& \textbf{AIDE}
& \textbf{HydraPrompt (\textbf{$\triangle$)}} \\
\midrule
\midrule 
ProGAN \cite{gao2019progan}  
& \begin{tabular}{@{}c@{}}56.9 \\ 0.1/99.7\end{tabular} 
& \begin{tabular}{@{}c@{}}55.6 \\ 13.7/87.1\end{tabular} 
& \begin{tabular}{@{}c@{}}57.0 \\ 0.0/99.8\end{tabular} 
& \begin{tabular}{@{}c@{}}58.9 \\ 4.8/99.7\end{tabular} 
& \begin{tabular}{@{}c@{}}57.1 \\ 0.1/99.9\end{tabular} 
& \begin{tabular}{@{}c@{}}57.2 \\ 3.2/97.8\end{tabular} 
& \begin{tabular}{@{}c@{}}58.2 \\ 3.3/99.5\end{tabular} 
& \begin{tabular}{@{}c@{}}53.8 \\ 1.8/92.8\end{tabular} 
& \begin{tabular}{@{}c@{}}57.3 \\ 2.2/98.7\end{tabular} 
& \begin{tabular}{@{}c@{}}58.4 \\ 5.0/98.5\end{tabular} 
& \begin{tabular}{@{}c@{}}\textbf{61.3} ($\textbf{\textcolor{red}{+2.4}}$) \\ 31.6/83.7\end{tabular} \\
\midrule
SD v1.4 \cite{ho2020denoising}  
& \begin{tabular}{@{}c@{}}60.1 \\ 8.9/98.6\end{tabular} 
& \begin{tabular}{@{}c@{}}56.9 \\ 1.4/98.6\end{tabular} 
& \begin{tabular}{@{}c@{}}57.1 \\ 0.0/99.9\end{tabular} 
& \begin{tabular}{@{}c@{}}61.0 \\ 17.7/93.5\end{tabular} 
& \begin{tabular}{@{}c@{}}55.6 \\ 0.6/97.0\end{tabular} 
& \begin{tabular}{@{}c@{}}55.6 \\ 75.0/44.1\end{tabular} 
& \begin{tabular}{@{}c@{}}59.7 \\ 11.9/95.7\end{tabular} 
& \begin{tabular}{@{}c@{}}56.3 \\ 3.1/96.4\end{tabular} 
& \begin{tabular}{@{}c@{}}59.1 \\ 2.4/100.0\end{tabular} 
& \begin{tabular}{@{}c@{}}62.6 \\ 20.3/94.4\end{tabular} 
& \begin{tabular}{@{}c@{}}\textbf{69.7} ($\textbf{\textcolor{red}{+7.1}}$) \\ 37.9/93.9\end{tabular} \\
\bottomrule
\end{tabular}
\caption{Accuracy comparisons with previous state-of-the-art methods on Chameleon \cite{yan2024sanity} dataset. The first row indicates accuracy \textbf{(Acc)} evaluated on Chameleon test dataset, and the second row gives the Acc for ``\textbf{fake image / real image}'' for detailed analysis. ``$\triangle$'' denotes the performance improvement compared with previous best method and are highlighted in red. The best results are highlighted in \textbf{bold}.}
\label{tab:your_label}
\end{table*}

\begin{table}[t] 
\centering
\footnotesize  
\vspace{-13pt} 
\renewcommand{\arraystretch}{1.0}
\newcolumntype{|}{!{\vline}}
\vspace{-5pt}
\begin{tabular}{
    >{\raggedright\arraybackslash}p{2.55cm} | 
    >{\centering\arraybackslash}p{0.88cm}
    >{\centering\arraybackslash}p{0.88cm}
    >{\centering\arraybackslash}p{0.88cm} |
    >{\centering\arraybackslash}p{0.88cm}
}
\toprule
\textbf{Methods} 
& {Facebook} 
& {Reddit} 
& {Twitter} 
& Mean \\
\midrule 
\midrule
CNNSpot \cite{wang2020cnn}  & 70.6 & 75.4 & 71.4 & 72.5 \\
PatchFor \cite{chai2020makes} & 77.1 & 87.8 & 81.6 & 82.2 \\
UnivFD \cite{ojha2023towards} & 78.4 & 80.8 & 78.1 & 79.1 \\
NPR \cite{tan2024rethinking} & 76.6 & 89.8 & 79.5 & 81.9 \\
LaDeDa \cite{cavia2024real} & 81.9 & 91.8 & 83.3 & 85.7 \\
HFMF \cite{mehta2025hfmf}  & 86.9 & 92.3 & 85.8 & 89.4 \\
\midrule
\rowcolor{Gray}HydraPrompt (Ours) & \textbf{95.2} & \textbf{95.3} & \textbf{97.3} & \textbf{95.9} \\
\rowcolor{Gray}\textbf{$\triangle$ Prev. Best} & $\textbf{\textcolor{red}{+8.3}}$ & $\textbf{\textcolor{red}{+3.0}}$ & $\textbf{\textcolor{red}{+11.5}}$ & $\textbf{\textcolor{red}{+6.5}}$ \\
\bottomrule
\end{tabular}
\caption{Accuracy comparisons with previous state-of-the-art methods on WildRF \cite{cavia2024real} dataset. All methods are trained on the WildRF training set and evaluated on the three subsets of the WildRF test datasets. We report the performance in the formulation of Acc. The best results are highlighted in \textbf{bold}, the gains are in red.}
\label{Table:WildRF dataset}
\vspace{-15pt} 
\end{table}

\subsection{Main Results}
\label{sec:Evaluation Analysis.}
\textbf{Comparison on UniversalFakeDetect benchmark}. We choose several previously popular methods, including \cite{wang2020cnn, durall2020watch, frank2020leveraging, qian2020thinking, tan2023learning, ojha2023towards, tan2024frequency, koutlis2024leveraging, liu2024forgery, tan2025c2p, yan2024orthogonal} as shown in Table~\ref{table:GAN datasets on UniFD} for comparison with our method. Compared with other prompt-based VLM approaches, such as (1) FatFormer \cite{liu2024forgery}, which designs a forgery-aware adapter and a text-guided interactor, our method achieves a cumulative performance improvement of 1.9$\%$ and 3.1$\%$ on GANs and Diffusion model datasets. Similarly, (2) C2P-CLIP \cite{tan2025c2p}, which injects category concepts into the image encoder through the text encoder, our method achieves gains of 2.7$\%$ and 2.9$\%$, respectively. Beyond prompt-based VLM approaches, such as RINE \cite{koutlis2024leveraging}, which exploits fine-grained visual cues as discriminative elements, our method achieves a cumulative improvement of 2.6$\%$ and 5.5$\%$, respectively. Effort \cite{yan2024orthogonal}, which explicitly constructs the orthogonal semantic and forgery subspaces via singular value decomposition (SVD), our method achieves an additional gain of 0.7$\%$ and 1.6$\%$, respectively. This performance on unseen generators supports the effectiveness and robustness of our framework. 

\noindent \textbf{Comparison on Chameleon dataset}. We choose several previously popular methods, including CNNSpot \cite{wang2020cnn}, FreDect \cite{frank2020leveraging}, Fusing \cite{ju2022fusing}, GramNet \cite{liu2020global}, LNP \cite{liu2022detecting}, UnivFD \cite{ojha2023towards}, DIRE \cite{wang2023dire}, PatchCraft \cite{zhong2023rich}, NPR \cite{tan2024rethinking}, AIDE \cite{yan2024sanity} for comparison with our method. Note that in the case of widespread post-processing artifacts and challenging test samples in Chameleon dataset, our method achieves an accuracy of over 60$\%$ under ProGAN training setting, achieving an improvement of 2.4$\%$ respectively. Under SD v1.4 training setting, our method still achieves improvements of 7.1$\%$ respectively, with an accuracy approaching 70$\%$.

\noindent \textbf{Comparison on WildRF dataset}. This is a dataset curated from several popular social networks, dedicated to evaluating model performance in online environments. We choose several previous popular methods, including \cite{wang2020cnn, chai2020makes, ojha2023towards, tan2024rethinking, cavia2024real, mehta2025hfmf} as shown in Table~\ref{Table:WildRF dataset}. It is noteworthy that in the online environments, our method still achieves over 90$\%$ accuracy under the WildRF training setting, with an improvement of 8.3$\%$, 3.0$\%$ and 11.5$\%$ on these three test datasets compared with previous best methods, respectively. 

\subsection{Analysis and Ablation Study}
\label{sec:Ablation Study and Analysis.}

We conduct the following analysis and ablation studies to comprehensively evaluate the effectiveness of each component of our framework. Unless specified, we report the mean of Acc and AP of the GANs and Diffusion model setting on UniversalFakeDetect \cite{ojha2023towards} benchmark.

\noindent \textbf{Robustness of a single set of prompts for authentic images.}
To analyze this, we conduct additional evaluations on diverse real datasets in Table~\ref{table:static prompt for authentic category}. The SID tasks primarily focus on fine-grained details. In this regard, real images possess a consistent distribution by the unified physics of real-world imaging. This physical invariance allows a static prompt to effectively anchor the real distribution, even across vast semantic variations.

\begin{figure}[t]
\centering
  \includegraphics[width=0.47\textwidth]{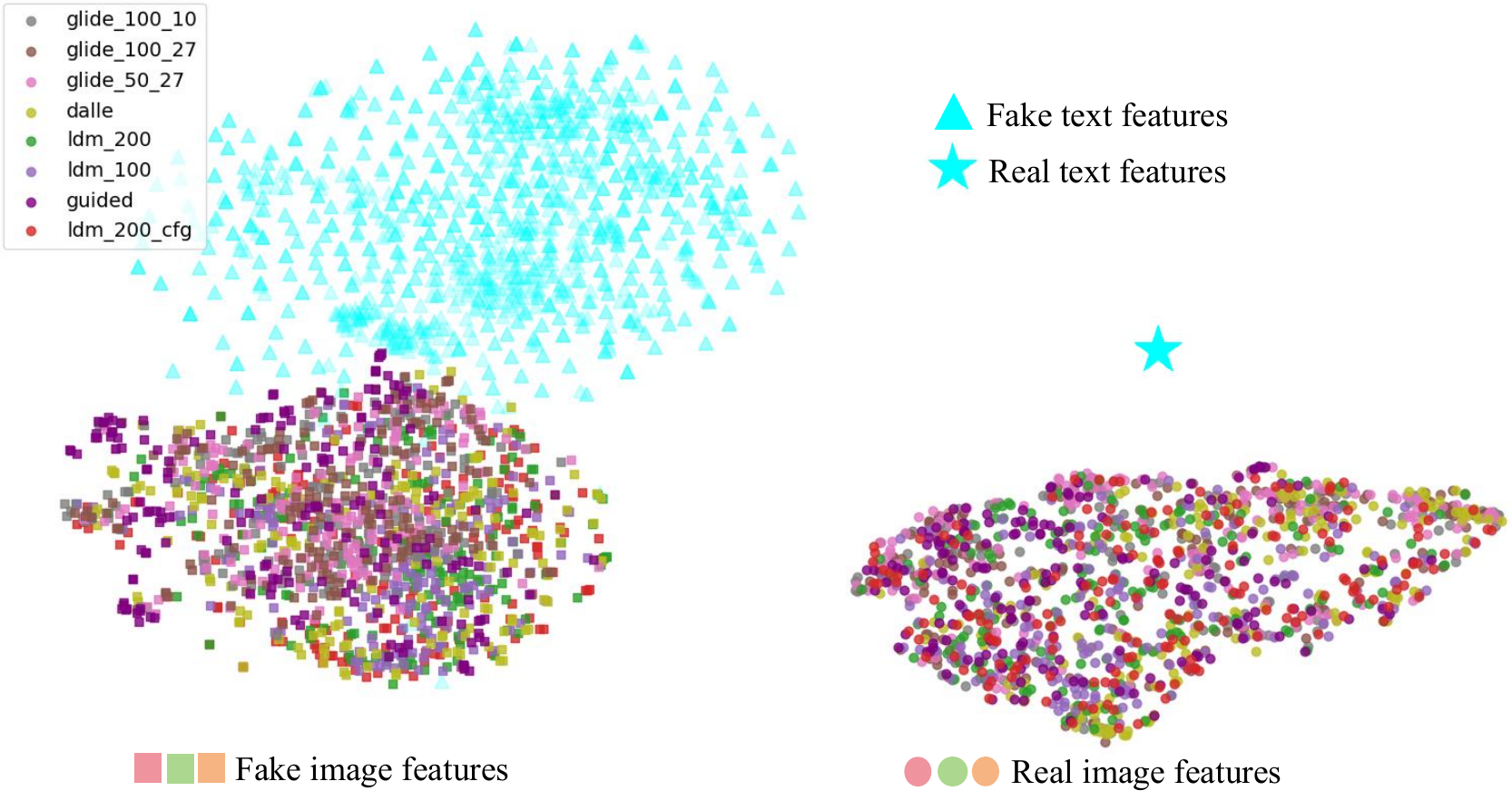}
  \caption{TSNE of Hydraprompt on \textbf{unseen} subsets from UniversalFakeDetect \cite{ojha2023towards} benchmark in cross-modal embedding. }
  \label{figure: Qualitative analysis} 
  \vspace{-10pt}
\end{figure}

\noindent \textbf{Robustness of adaptive prompts for fake images.} 
Qualitatively, we selected the unseen subsets from the UniversalFakeDetect\cite{ojha2023towards} benchmark in Figure~\ref{figure: Qualitative analysis}. It shows the overall fake text and image features are consistently more closely aligned, while maintaining a distinct decision boundary from the real class distribution. This suggests that the distribution of adaptive prompts is highly regularized under the CSC objective. 

\noindent \textbf{Performance compared with similar prompt-tuning method.}
We strictly reproduce the CoOp and CoCoOp algorithms and conduct the following experiments in Table~\ref{table:compared with similar prompt tuning method}. The SID tasks primarily focus on fine-grained details. In this regard, SID is intrinsically asymmetric, distinguishing it from the general semantic tasks associated with CoCoOp. An adaptive prompt for the real class may disrupt the unified physical distribution of real-world imaging.

\begin{table}[t]
\centering
\footnotesize
\renewcommand{\arraystretch}{1.0}
\begin{tabular}{
  >{\raggedright\arraybackslash}p{3.5cm} |
  >{\centering\arraybackslash}p{1.0cm}
  >{\centering\arraybackslash}p{1.0cm}
  >{\centering\arraybackslash}p{1.0cm}
  >{\centering\arraybackslash}p{1.3cm}
  >{\centering\arraybackslash}p{0.8cm}
}
\toprule
\multicolumn{1}{c|}{Metric} & {UniFD\cite{ojha2023towards}} & {WildRF\cite{cavia2024real}} & {LOKI\cite{girase2021loki}} & {MSCOCO\cite{lin2014microsoft}} & \cellcolor{gray!20}{Avg.} \\ 
\cline{1-6}
\multicolumn{1}{c|}{Acc} 
& 98.8
& 95.7
& 91.6
& 98.9
& \cellcolor{gray!20}{96.3} \\ 
\bottomrule
\end{tabular}
\caption{\textbf{Quantitative analysis} $\textbf{(\%)}$ of stability for the static prompt for high-variance authentic images. Specifically, LOKI\cite{girase2021loki} contains proprietary real images such as medicine and remote sensing, while MSCOCO\cite{lin2014microsoft} covers the distribution of the vast majority of real-world imagery.}
\label{table:static prompt for authentic category}
\vspace{-15pt} 
\end{table}

\begin{table}[t]
\centering
\footnotesize
\renewcommand{\arraystretch}{1.0}
\begin{tabular}{
  >{\raggedright\arraybackslash}p{3.5cm} |
  >{\centering\arraybackslash}p{0.8cm}
  >{\centering\arraybackslash}p{0.8cm}
  >{\centering\arraybackslash}p{1.1cm}
}
\toprule
\multicolumn{1}{c|}{Method} & {Acc} & {AP} & {Avg.} \\ 
\cline{1-4}
\multicolumn{1}{c|}{CoOp\cite{zhou2022learning}} 
& 92.1
& 97.5
& 94.8 \\ 
\multicolumn{1}{c|}{CoCoOp\cite{zhou2022conditional}} 
& 90.8 
& 96.0 
& 93.4  \\ 
\rowcolor{Gray}\multicolumn{1}{c|}{HydraPrompt} 
& \textbf{94.7} 
& \textbf{98.2} 
& \textbf{96.5}  \\ 
\cline{1-4}
\rowcolor{Gray}\multicolumn{1}{c|}{$\triangle$ CoOp baseline} 
& (\textcolor{red}{\textbf{+2.6}}) 
& (\textcolor{red}{\textbf{+0.7}}) 
& (\textcolor{red}{\textbf{+1.7}}) \\ 
\bottomrule
\end{tabular}
\caption{Performance comparison $\textbf{(\%)}$ with similar prompt-tuning method. It can be observed that in SID task, HydraPrompt's performance is significantly superior. Therefore, the asymmetric design of HydraPrompt reflects a principled modeling choice rather than an empirical refinement of existing prompt-tuning strategies.}
\label{table:compared with similar prompt tuning method}
\vspace{-20pt} 
\end{table}

\begin{table}[t]
\centering
\footnotesize
\renewcommand{\arraystretch}{1.0}
\begin{tabular}{
  >{\raggedright\arraybackslash}p{1.5cm} |
  >{\raggedright\arraybackslash}p{0.2cm} |
  >{\centering\arraybackslash}p{0.6cm}
  >{\centering\arraybackslash}p{0.8cm}
  >{\centering\arraybackslash}p{1.0cm}
  >{\centering\arraybackslash}p{1.9cm}
}
\toprule
\multicolumn{1}{c|}{Metrics} & {bs} & {CoOp} & {CoCoOp} & {FatFormer\cite{liu2024forgery}}  & HydraPrompt \\ 
\cline{1-6}
\multicolumn{1}{c|}{\multirow{2}{*}{FLOPs(G)}}
& 1
& 94.55
& 94.55 
& 127.95
& 94.56  \\ 
\cline{2-6}
\multicolumn{1}{c|}{} 
& 8
& 664.76
& 756.44
& \cellcolor{gray!20}{1023.63}
& \cellcolor{gray!20}{710.61}(\textcolor{mgreen}{\textbf{-313.02}}) \\ 
\cline{1-6}
\multicolumn{1}{c|}{\multirow{2}{*}{Latency(ms)}}
& 1
& 28.98
& 31.96 
& 125.37
& 32.42 \\ 
\cline{2-6}
\multicolumn{1}{c|}{} 
& 8
& 59.63
& 65.19
& \cellcolor{gray!20}{216.76}
& \cellcolor{gray!20}{63.17} (\textcolor{mgreen}{\textbf{-153.59}}) \\ 
\bottomrule
\end{tabular}
\caption{\textbf{Quantitative analysis} of the compuatational overhead of HydraPrompt and several previous popular methods. ``bs'' refers to batch size.}
\label{table:computational overhead}
\vspace{-15pt} 
\end{table}

\begin{table}[t]
\centering
\footnotesize
\renewcommand{\arraystretch}{1.0}
\begin{tabular}{
  >{\raggedright\arraybackslash}p{3.5cm} |
  >{\centering\arraybackslash}p{1.8cm} |
  >{\centering\arraybackslash}p{0.8cm}
  >{\centering\arraybackslash}p{0.8cm}
  >{\centering\arraybackslash}p{1.2cm}
}
\toprule
\multicolumn{1}{c|}{visual cues} & {Type} & {Acc} & {AP} & {Avg.} \\ 
\midrule
\rowcolor{Gray}\multicolumn{1}{c|}{Layer 1} 
& Low-level
& \textbf{96.9} 
& \textbf{99.6}
& \textbf{98.2} \\ 
\multicolumn{1}{c|}{Layer 12} 
& Mid-level
& 94.2
& 99.0
& 96.6 (\textcolor{mgreen}{\textbf{-1.6}}) \\ 
\multicolumn{1}{c|}{Layer 24} 
& High-level
& 91.4
& 98.0 
& 94.7 (\textcolor{mgreen}{\textbf{-3.5}}) \\ 
\multicolumn{1}{c|}{LDR-Net \cite{chen2025ldr}} 
& Frequency
& 95.3
& 97.8
& 96.5 (\textcolor{mgreen}{\textbf{-1.7}}) \\ 
\multicolumn{1}{c|}{RINE \cite{koutlis2024leveraging}} 
& Ensemble
& 95.4
& 98.8  
& 97.1 (\textcolor{mgreen}{\textbf{-1.1}}) \\ 
\bottomrule
\end{tabular}
\caption{\textbf{Quantitative analysis} $\textbf{(\%)}$ of fine-grained visual cues in APA Module. We adopt Layer 1 setting in HydraPrompt.}
\label{table:Quantitative analysis:fine-grained visual cues}
\vspace{-15pt} 
\end{table}

\noindent \textbf{Computational overhead}.We calculate the computational overhead of HydraPrompt and several previous popular methods in Table~\ref{table:computational overhead}, and the results show that HydraPrompt has relatively higher computational efficiency.

\begin{table*}[t]
\centering
\footnotesize 
\renewcommand{\arraystretch}{1.0}
\vspace{-5pt} 
\begin{subtable}[t]{0.35\textwidth}
\centering
\begin{tabular}{
  >{\centering\arraybackslash}p{0.5cm} 
  >{\centering\arraybackslash}p{0.5cm} |
  >{\centering\arraybackslash}p{0.5cm} 
  >{\centering\arraybackslash}p{0.5cm} |
  >{\centering\arraybackslash}p{0.4cm} 
  >{\centering\arraybackslash}p{0.4cm} 
  >{\centering\arraybackslash}p{0.4cm}  
}
\toprule
\multicolumn{2}{c|}{Real} 
&\multicolumn{2}{c|}{Fake} 
& \multirow{2}{*}{Acc}  
&\multirow{2}{*}{AP} 
& \multirow{2}{*}{Avg.} \\
Ada.
&Static
&Ada.
&Static
& & & \\
\midrule
$\checkmark$
& 
& 
& $\checkmark$
& 89.2
& 96.1
& 92.6 \\ 
$\checkmark$
& 
& $\checkmark$
& 
& 90.2
& 95.7
& 92.9 \\ 
& $\checkmark$
& 
& $\checkmark$
& 91.5
& 96.7 
& 94.1 \\ 
\rowcolor{Gray}& $\checkmark$
& $\checkmark$
& 
& \textbf{96.9} 
& \textbf{99.6}
& \textbf{98.2} \\ 
\midrule
\rowcolor{Gray}\multicolumn{4}{c}{\hspace{-5pt}\textbf{$\triangle$ Baseline}}
& $\textcolor{red}{\textbf{+5.4}}$
& $\textcolor{red}{\textbf{+2.9}}$
& $\textcolor{red}{\textbf{+4.1}}$ \\ 
\bottomrule
\end{tabular}
\caption{\textbf{Ablation $\textbf{(\%)}$} on prompts design in APA Module. ``Ada.'' denotes Adaptive. The baseline is ``Real-Static'' and ``Fake-Static'' setting.}
\label{table:APA}
\end{subtable}
\hfill
\begin{subtable}[t]{0.33\textwidth}
\centering
\begin{tabular}{
  >{\raggedright\arraybackslash}p{0.35cm} 
  >{\centering\arraybackslash}p{0.36cm} |
  >{\centering\arraybackslash}p{0.35cm} 
  >{\centering\arraybackslash}p{0.36cm} |
  >{\centering\arraybackslash}p{0.35cm} 
  >{\centering\arraybackslash}p{0.35cm} 
  >{\centering\arraybackslash}p{0.35cm} 
}
\toprule
\multicolumn{2}{c|}{Real} 
&\multicolumn{2}{c|}{Fake} 
& \multirow{2}{*}{Acc}  
&\multirow{2}{*}{AP} 
& \multirow{2}{*}{Avg.} \\
Ind.
&Clus.
&Ind.
&Clus.
& \\ 
\midrule
$\checkmark$
& 
& 
& $\checkmark$
& 92.9
& 99.3
& 96.1 \\ 
$\checkmark$
& 
& $\checkmark$
& 
& 90.2
& 95.4 
& 92.8 \\ 
& $\checkmark$
&
& $\checkmark$
& 94.7
& 98.2
& 96.5 \\ 
\rowcolor{Gray}& $\checkmark$
& $\checkmark$
& 
& \textbf{96.9} 
& \textbf{99.6}
& \textbf{98.2} \\ 
\midrule
\rowcolor{Gray}\multicolumn{4}{c}{\textbf{$\triangle$ Baseline}}
& $\textcolor{red}{\textbf{+2.2}}$
& $\textcolor{red}{\textbf{+1.4}}$
& $\textcolor{red}{\textbf{+1.7}}$ \\ 
\bottomrule
\end{tabular}
\caption{\textbf{Ablation $\textbf{(\%)}$} of $\mathbf{M}_{i j}$ on CSC Objective. ``Ind.'', ``Clus.'' denotes Individual and Cluster. The baseline is ``Real-Clus.'' and ``Fake-Clus.'' setting.}
\label{table:CSC}
\end{subtable}
\hfill
\hspace{-15pt}
\begin{subtable}[t]{0.31\textwidth}
\centering
\renewcommand{\arraystretch}{1.5}
\captionsetup{width=0.8\textwidth} 

\begin{tabular}{
  >{\centering\arraybackslash}p{0.55cm} 
  >{\centering\arraybackslash}p{0.55cm} |
  >{\centering\arraybackslash}p{0.45cm} 
  >{\centering\arraybackslash}p{0.45cm} 
  >{\centering\arraybackslash}p{0.45cm} 
}
\toprule
\multicolumn{2}{c|}{$\mathcal{L}_{\text{sc}}$} 
& \multirow{2}{*}{Acc} 
& \multirow{2}{*}{AP} 
& \multirow{2}{*}{Avg.} \\
\cline{1-2}
 \rule{0pt}{9pt} text 
& image
& \\
\midrule
$\checkmark$ & $\times$ & 92.2 & 96.8 & 94.5 \\
$\times$ & $\checkmark$ & 94.1 & 98.0 & 96.1 \\
\rowcolor{Gray}$\checkmark$ & $\checkmark$ & \textbf{96.9} & \textbf{99.6} & \textbf{98.2} \\
\bottomrule
\end{tabular}
\caption{\textbf{Ablation $\textbf{(\%)}$} on $\mathcal{L}_{\text{sc}}$ design on CSC Objective. We adopt text and image branches setting. }
\label{table:SC loss}
\end{subtable}
\caption{\textbf{Ablation $\textbf{(\%)}$} experiments for HydraPrompt. We choose three ablation stydies below, including prompt design in APA Module (a), $\mathbf{M}_{i j}$ on CSC Objective (b), and $\mathcal{L}_{\text{sc}}$ design on CSC Objective (c). Default settings are marked in gray.}
\end{table*}

\noindent \textbf{Quantitative analysis of fine-grained visual cues in APA Module.}
In Table \ref{table:Quantitative analysis:fine-grained visual cues}, we compare different strategies for extracting fine-grained visual cues, including shallow (layer 1), middle (layer 12), and deep (layer 24) features from image encoder, the hand-crafted local cues based on neighboring pixel variations proposed by LDR-Net \cite{chen2025ldr} and the approach of RINE \cite{koutlis2024leveraging}, which concatenates CLS tokens from all encoder layers by a fully connected layer. The experimental results demonstrate that under our APA Module and CSC Objective, using features from the first layer of image encoder achieves the best performance.

\noindent \textbf{Ablation on the APA designs.}
In Table~\ref{table:APA}, we compare the impact of different prompt configurations on model performance. ``Ada.'' refers to sample-adaptive prompt, which incorporates fine-grained visual cues, while static denotes a single set of learnable prompts. The baseline (\ie, ``Real-Static'' and ``Fake-Static'') achieves 91.5$\%$ Acc and 96.7$\%$ AP. Compared to this baseline, the incorporation of our APA Module significantly improves the performance, with gains of 5.4$\%$ in Acc and 2.9$\%$ in AP. Note that \textit{when both the real and fake prompts are set to ``Ada.'', the model fail to yield a relative performance improvement}. As shown in Figure~\ref{figure:observation}, an asymmetric phenomenon exists in the SID task, leading these symmetric text prompts, although dynamic, to fail in providing an effective adaptive category center.

\noindent \textbf{Ablation on CSC objective designs.}
In Table~\ref{table:CSC}, we compare different combinations of the CSC objective. ``Ind.'' denotes that each feature is regarded as an individual by pushing the feature apart from one another. ``Clus.'' denotes all features are regarded as a single cluster by pulling them together. Under our APA Module, the baseline (\ie, ``Real-Cluster'' and ``Fake-Cluster'') achieves 94.7$\%$ Acc and 98.2$\%$ AP. On this basis, our proposed objective further improves the performance by 2.2$\%$ in Acc and 1.4$\%$ in AP. In Table~\ref{table:SC loss}, we compare different combinations of the $\mathcal{L}_{\text{sc}}$. The experiments show that applying the $\mathcal{L}_{\text{sc}}$ to different branches has a certain impact on the model's performance, and the optimal performance is achieved when the $\mathcal{L}_{\text{sc}}$ is applied to both the text and image branches simultaneously.

\begin{figure}[t]
\centering
\vspace{-10pt} 
  \includegraphics[width=0.47\textwidth]{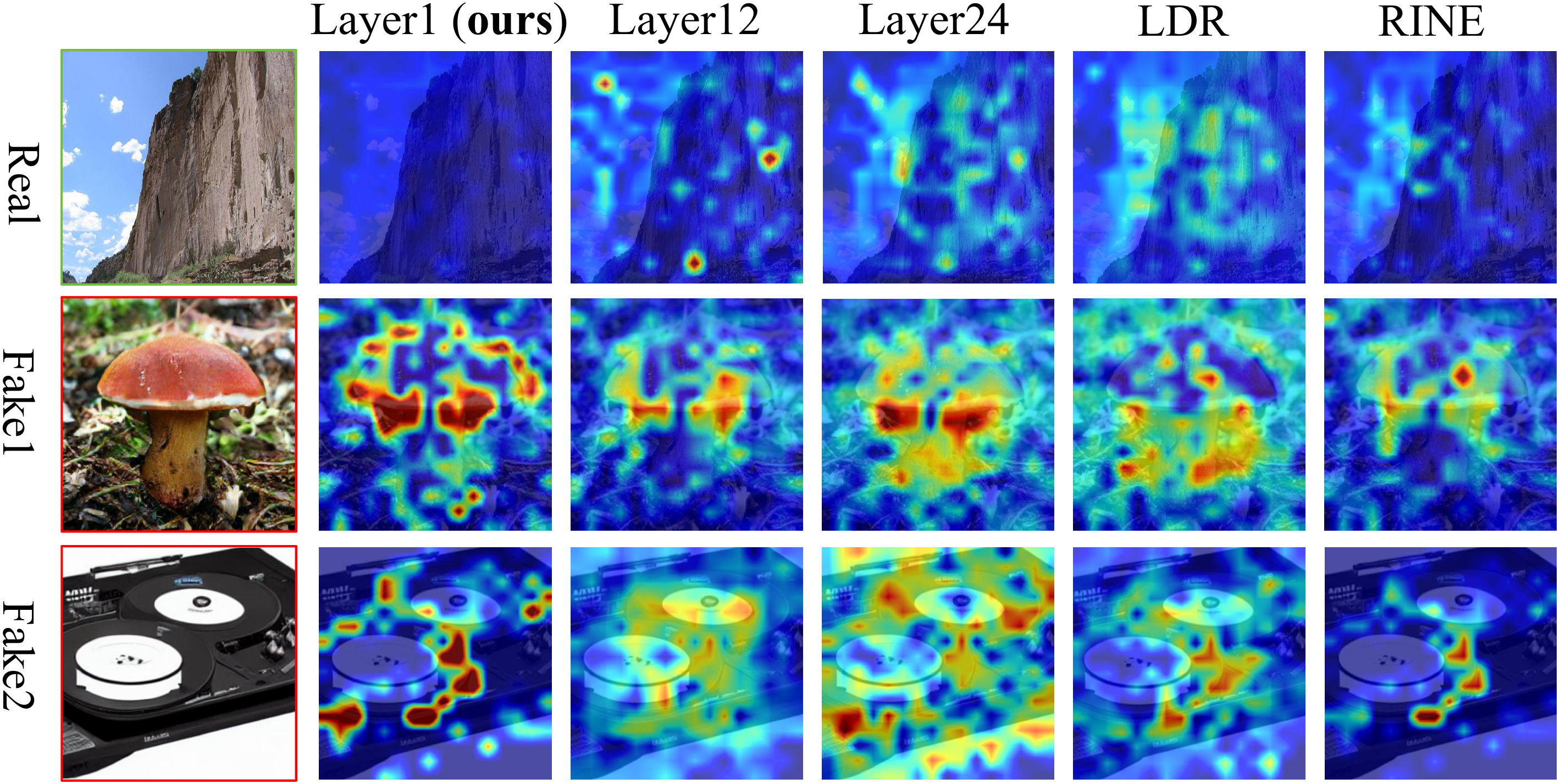} 
  \vspace{-5pt} 
  \caption{\textbf{Qualitative analysis} of fine-grained visual cues by gradient norm \cite{selvaraju2017grad} visualization, which is trained with five different designs in APA, including the shallow layer (Layer 1), middle layer (layer 12), deep layer (Layer 24) of image encoder, along with two previous popular methods (\ie, LDR-net \cite{chen2025ldr}), RINE \cite{koutlis2024leveraging}. We select the gradient norm based on the logit of fake class, therefore, a higher value indicates greater attention to forgery artifacts. The layer 1 setting is adopted in our HydraPrompt.}
  \label{figure: Qualitative analysis: fine-grained cues} 
\vspace{-20pt} 
\end{figure}

\subsection{ Further Analysis }
\label{sec: Further Analysis.}

\noindent \textbf{Qualitative analysis of fine-grained visual cues in APA Module}. We visualize the gradient norm \cite{selvaraju2017grad} maps of forgery traces under different strategies in Figure~\ref{figure: Qualitative analysis: fine-grained cues}. Our framework adopts the strategy of Layer 1. (1) For authentic category, the models bulit with Layer 1 exhibit less attention to forgery traces, whereas others exhibit unreasonable attention patterns. (2) For synthetic category, the models bulit with shallow layer features (Layer 1) focus more on key regions, such as the structure and contours of the mushroom in ``Fake 1'', and the shape and edges of the disc in ``Fake 2''. In contrast, models built with deeper features (Layer 24) and frequency domain fingerprints (LDR-Net \cite{chen2025ldr}) exhibit more dispersed attention areas and are relatively weaker in capturing crucial information. Based on this, we conclude that shallow layers of image encoder capture fine-grained details, which are more crucial in SID task, thereby enabling the model to learn more effective representations.

\begin{figure}[t]
\centering
  \vspace{-10pt}
  \includegraphics[width=0.475\textwidth]{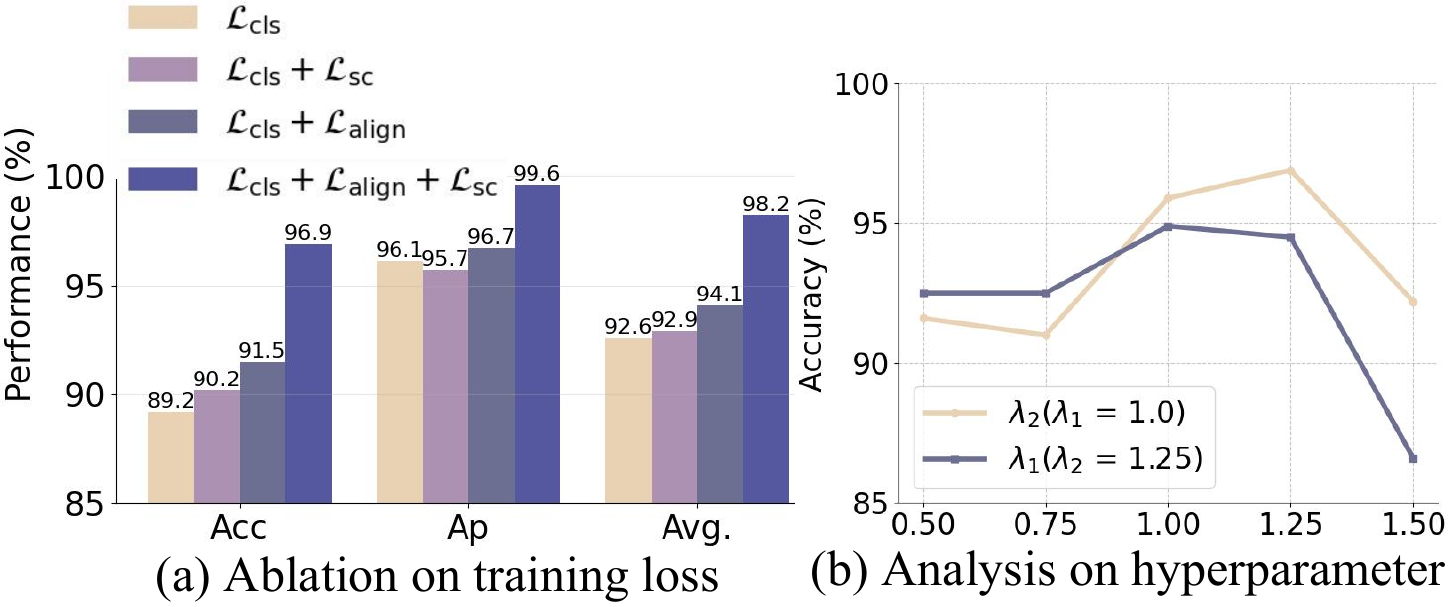} 
  \caption{(a) \textbf{Ablation $\textbf{(\%)}$} on training loss design, including $\mathcal{L}_{\text{cls}}$, $\mathcal{L}_{\text{sc}}$, $\mathcal{L}_{\text{align}}$. (b) \textbf{Analysis} of hyperparameter.}
  \label{figure: Ablation on training loss} 
\vspace{-10pt} 
\end{figure}

\noindent \textbf{Ablation on training loss}.
We compare different combinations of training loss in Figure~\ref{figure: Ablation on training loss} (a). The results demonstrate that the model achieves optimal performance when both $\mathcal{L}_{\text{align}}$ and $\mathcal{L}_{\text{sc}}$ are incorporated. It can be observed that adding only one of them has a limited impact on performance, as $\mathcal{L}_{\text{sc}}$ focues on the intra-modal features alignment, while $\mathcal{L}_{\text{align}}$ operates on the cross-modal features alignment. Therefore, the model can learn more effective representations when these two constraints work collaboratively.

\noindent \textbf{Sensitivity analysis of the parameters $\lambda_1$ and $\lambda_2$ on training loss}. We select different combinations of $\lambda_1$ and $\lambda_2$ on validation sets of UniversalFakeDetect \cite{ojha2023towards} benchmark in Figure~\ref{figure: Ablation on training loss} (b), it can be observed that these hyperparameters also have a noticeable impact on the experimental results. It shows the optimal performance can be achieved when $\lambda_1$ is set to 1.0 and $\lambda_2$ is set to 1.25.

\section{Conclusion}
\label{sec:Conclusion}

In this paper, we propose HydraPrompt, an adaptive and asymmetric prompting framework that achieves adaptive category centers during inference. It employs an APA module for constructing sample-adaptive prompts, supported by a CSC objective to boost adaptation ability. 
Experiments on popular SID benchmarks from distinct perspectives demonstrate the state-of-the-art performance of HydraPrompt.


\bibliographystyle{ACM-Reference-Format}
\bibliography{main}










\end{document}